# Reinforcement learning on the EPRI AI-ccelerating Unit Commitment benchmark: diagnosing over-commitment and guiding MILP with learned policies

Maharani Devira Pramita[a], Aidan O'Sullivan[a,*]

[a] UCL Energy Institute, Bartlett School of Environment, Energy and Resources, University College London, Central House, 14 Upper Woburn Place, London WC1H 0NN, United Kingdom

* Corresponding author. E-mail address: aidan.osullivan@ucl.ac.uk

## Highlights

- First RL study on the EPRI benchmark: agents commit twice the optimal fleet.
- Spare units insure against exploration: injected noise enlarges the fleet.
- Imitation and conservative fine-tuning give the lowest penalised cost tested.
- Fixing the fine-tuned policy's confident decisions speeds the MILP 3.8–8.5-fold.
- A scope-matched optimum and a simple rule are needed to judge learned schedulers.

## Abstract

Reinforcement learning (RL) is increasingly proposed as a fast solver for unit commitment (UC), but standalone agents remain far more expensive than mixed-integer linear programming (MILP). In the first study of RL on the 51-generator, 72-hour benchmark of the Electric Power Research Institute (EPRI) AI-ccelerating Unit Commitment competition, we trace this gap to the agents' own exploration, close most of it, and show that the resulting policy can accelerate an exact solver. In every configuration tested, Proximal Policy Optimisation (PPO) and an action-branching Deep Q-Network (DQN) trained from scratch committed about twice the units of a thermal-only MILP optimum, at 39–64% higher cost, whereas a merit-order rule came within 1.6–7% (10–24% when unserved energy is priced). When each unit's decision is sampled independently, spare units insure against random de-commitments. Consistent with this, injecting exploration noise during fine-tuning enlarged a lean policy's fleet, a reward without shaping terms did not remove the over-commitment, and under their training noise the agents outperformed the rule subjected to comparable noise. Imitating the MILP solutions and fine-tuning conservatively brought PPO within 2.6% of the optimum, and these policies had the lowest penalised cost of any evaluated (about 5%). Fixing the fine-tuned policy's confident decisions made the MILP 3.8–8.5 times faster within 0.10–0.25% of the best solution, with a smaller worst case than an imitation-only policy fixing as many variables (1.1% against 8.9%). Because dispatch method and benchmark scope alone shift the reported optimality ratio by up to 3.6% and 9%, we propose a reporting protocol.

## Nomenclature

| Symbol | Description |
|---|---|
| $i \in \mathcal{G}$ | Thermal generator index; $\lvert\mathcal{G}\rvert = 51$ |
| $t \in \{1, \dots, T\}$ | Hourly time step; $T = 72$ |
| $u_{i,t}, s_{i,t}, w_{i,t}$ | Commitment status, start-up and shut-down indicators (binary) |
| $p_{i,t}$ | Dispatched output (MW) |
| $\underline{P}_i, \overline{P}_i$ | Minimum stable and maximum output (MW) |
| $C_i(p)$ | Convex piecewise-linear production cost of unit $i$, including no-load cost ($/h) |
| $c_i$ | First-segment marginal cost of unit $i$ ($/MWh) |
| $R_i^{\text{up}}$ | Ramp-up limit of unit $i$ (MW/h) |
| $D_t^{\text{net}}$ | Net demand: demand minus wind and solar output (MW) |
| $N_t$ | Number of committed generators at hour $t$ |
| $C_i^{\text{su}}$ | Start-up cost ($) |
| $\lambda^{\text{uns}}, \lambda^{\text{sp}}$ | Unserved-energy and spill penalties ($/MWh) |
| $E_t^{\text{uns}}, E_t^{\text{sp}}$ | Unserved energy and spill at hour $t$ (MWh) |
| $r_t$ | Reward at hour $t$ |
| $F_t, S_t$ | Fuel and start-up cost at hour $t$ in the training reward ($) |
| $B_t$ | Optional per-hour baseline cost in the reward ($) |
| $\varepsilon$ | Per-unit exploration noise: probability that a free unit's executed decision differs from the greedy decision |
| $m$ | Capacity margin of the merit-order rule |
| $k$ | Number of units in the static-fleet control |
| $\rho$ | Within-episode correlation of $N_t$ with $D_t^{\text{net}}$ |
| $n \in \mathcal{N}$ | Test instance; $\lvert\mathcal{N}\rvert = 100$ |
| O.R. | Optimality ratio |

# 1. Introduction

This section sets out why the study was undertaken and what it contributes. Section 1.1 describes the computational burden of unit commitment and the gap between standalone RL and MILP, and Section 1.2 states the contributions.

## 1.1. Motivation

The unit commitment (UC) problem determines which thermal generating units to bring online over a multi-hour horizon so that demand is met at minimum cost, subject to minimum stable output, ramping, and minimum up- and down-time constraints [1,2]. The binary commitment decisions are coupled across units and across time, and the problem is strongly NP-hard [3]. Mixed-integer linear programming (MILP) is the industry standard, and tight formulations now solve large instances to small optimality gaps [4,5]. Its limitation is computational: each scenario requires a new combinatorial solve, and variable renewable generation increases how often schedules must be recomputed [6,7].

This computational burden has motivated a growing literature on learned approximate schedulers that move effort into an offline training phase and produce schedules in seconds at inference [8–10]. Reinforcement learning (RL) is a natural fit because commitment is sequential: a decision taken now constrains the choices available in later hours [11,12].

Model-free RL has been applied to UC at increasing scale [13–18], and hybrid methods that combine a learned policy with search or optimisation achieve cost close to MILP [19–22].

Standalone model-free policies remain more expensive than mathematical programming, and the gap is usually attributed to the intrinsic difficulty of coordinating many binary decisions [20,22]. Understanding why they fall short is the step that would let them close the gap. Learned schedulers, whether trained by RL or supervised learning, are usually evaluated by an optimality ratio against a MILP benchmark and a feasibility or fulfilment rate [9,13,16,17,20,21], which shows how far a schedule is from the benchmark but not what kind of schedule it is. Two further diagnostics would add to these: whether a policy has optimised its own training reward, and whether a simple rule would do as well. Both echo a broader literature on evaluation practice in deep RL, which has shown that conclusions drawn from few random seeds and point estimates are frequently unreliable [23–25].

### 1.2. Contributions

This paper presents, to the authors' knowledge, the first study of RL on the Electric Power Research Institute (EPRI) AI-ccelerating Unit Commitment benchmark [9,10,26]. It explains why standalone RL over-commits on this benchmark, a single-bus representation of the Irish test system with 51 thermal units and a 72-hour horizon, and shows how most of the gap to the optimum can be closed. The main contributions of this paper are that it:

1. establishes the first RL baselines on the benchmark, comparing Proximal Policy Optimisation (PPO) and an action-branching Deep Q-Network (DQN) with three references (a scope-matched MILP optimum, a merit-order rule and a static-fleet control) on cost and on the agents' own training reward, and finding that agents trained from scratch commit about twice the optimal fleet in every reward, masking and hyperparameter setting tested;
2. identifies a mechanism for the over-commitment and tests it directly: with factorised per-unit actions, spare capacity is reward-optimal insurance against the agent's own exploration. Injecting noise into a lean policy during fine-tuning enlarges its fleet, a reward without shaping terms leaves the over-commitment in place, the agents outperform the rule under comparable noise, and fine-tuned DQN's fleet increases with its exploration rate;
3. demonstrates that a lean PPO policy initialised by imitation drifts back towards over-commitment under standard but not conservative fine-tuning, and that DQN shows an additional, noise-independent drift;
4. finds that imitation of the MILP solutions followed by conservative fine-tuning, with a reward equal to the MILP objective and a look-ahead dispatcher, gives the lowest penalised cost of any policy evaluated;
5. shows that the fine-tuned policy can accelerate an exact solver: fixing its confident decisions gives a several-fold speed-up at a small cost, with a smaller worst case than an imitation-only policy fixing as many variables and a much smaller error than a merit-order heuristic;
6. quantifies how dispatch, unserved energy and benchmark scope change the reported optimality ratio, and sets out a reporting protocol for learned UC schedulers.

## 2. Literature review

This section reviews the work on which the study builds. Section 2.1 covers optimisation-based UC and the learned methods that support a MILP solver, Section 2.2 reinforcement learning for UC, and Section 2.3 imitation learning and fine-tuning. Section 2.4 identifies the gap this paper addresses.

### 2.1. Optimisation-based unit commitment

UC combines binary commitment with continuous dispatch decisions and inter-temporal constraints [1]. Lagrangian relaxation, dynamic programming and metaheuristics were historically prominent [2,27], but MILP became dominant once solvers and formulations matured, because it represents constraints exactly and provides optimality bounds [4,5]. Its persistent limitation is not solution quality but the cost of repeated solves, which has motivated learning methods that predict commitments or warm-start the solver [8–10], a use of learned predictions to fix variables that is also established for general mixed-integer programmes [28]. Two such methods have been reported on the EPRI benchmark, both supervised and both retaining a MILP solver at inference. The competition winner predicts commitment probabilities with a nearest-neighbour classifier and fixes confidently predicted variables, with generator-specific thresholds tuned on validation instances, reporting a mean optimality gap of 0.48% at a mean speed-up of 20.8 times [10]. The second combines a Transformer-based commitment predictor with deterministic feasibility repair and a warm-started MILP; the raw prediction is feasible in 31.6% of cases, while the full pipeline reaches 100% feasibility at a median optimality ratio of 1.0007 [9].

### 2.2. Reinforcement learning for unit commitment

Dalal and Mannor [13] formulated UC as a Markov decision process on a 12-generator system and obtained schedules faster than conventional search. Subsequent work applied tabular and distributed Q-learning [14,17], fuzzy RL [15], deep policy-gradient methods under load and wind uncertainty [16], PPO with separate dispatch optimisation [18], and quantum RL within a two-stage framework [29]. Reviews of RL in power systems identify scalability, constraint satisfaction and evaluation against optimisation baselines as recurring open problems [12,30,31].

Because standalone policies remain more expensive than MILP and the gap widens with system size [22], higher-performing approaches restrict or refine the learned policy: guided tree search over a policy-pruned action space [19,20,32], or optimisation-assisted candidate actions [21]. Related supervised approaches replace the RL policy with a commitment predictor that warm-starts or fixes variables in a MILP solver [9,10]; to our knowledge, whether an RL policy can serve the same purpose has not been tested. Most RL studies of UC consider 5 to 30 generators [13,20,21]; the 51-unit EPRI system is larger than most.

Because the joint action space grows as $2^{|\mathcal{G}|}$, standalone agents usually factorise the action into per-generator decisions that share a system reward, either as independent Bernoulli outputs of a policy or as separate value branches [33]. This resembles independent multi-agent learning [34] and value decomposition [35], but additivity does not hold in UC, because the value of committing one unit depends on which others are online. Independent learners in cooperative settings are known to prefer actions that are robust to their teammates' exploration over jointly optimal ones, a pathology termed relative overgeneralisation [36]. Masking invalid actions before sampling, rather than projecting them afterwards, yields a valid policy gradient [37,38]. Reward shaping preserves the optimal policy only when the shaping terms take a specific potential-based form [39,40].

That exploration shapes what an agent learns is well known in principle. In the cliff-walking example of Sutton and Barto [11], on-policy SARSA learns a longer, safer path than Q-learning because it evaluates the policy it follows, including its exploratory steps, rather than the greedy policy. Joint or autoregressive action heads, which condition each sub-action on those already chosen [41,42], are one way to avoid independent sampling across action dimensions. What has not been examined is how strongly this effect shapes the schedules of factorised RL agents for UC, where many binary decisions jointly secure a heavily penalised constraint.

### 2.3. Imitation and fine-tuning

When a competent policy or expert is available, behaviour cloning trains a policy by supervised learning on expert decisions [43]. Cloned policies are vulnerable to compounding errors once they leave the states the expert visited, which interactive methods such as Dataset Aggregation (DAgger) address by querying the expert on the learner's own states [44]. Demonstrations can also initialise RL: DQN from demonstrations combines temporal-difference learning with a large-margin loss on expert actions [45], and policy-gradient methods can be pre-trained on demonstrations and then fine-tuned [46,47]. A recurring observation is that online fine-tuning can erode the initial policy before the value estimates are reliable, which motivates conservative or constrained updates in offline-to-online RL [48–50]. In UC, the optimal MILP solutions supplied with a benchmark are a natural source of demonstrations, and the learned predictors that warm-start MILP solvers [8–10] are imitation learners in this sense.

### 2.4. Research gap

RL-for-UC studies usually report aggregate cost and feasibility against a MILP benchmark. Three diagnostics are not standard in this literature: comparison with a simple rule and with an optimum solved for exactly the problem the agent faces, a test of whether the agent has optimised its own training objective, and an explanation of the characteristic behaviour of policies that fall short of the benchmark. Published work on the EPRI benchmark has so far combined supervised prediction with a MILP solver [9,10]. This paper provides the first RL study on it, applies all three diagnostics, identifies the role of exploration in the gap, tests imitation-based remedies against the same optimum, and tests whether the resulting policy can guide an exact solver.

## 3. Problem setup

Here we provide an overview of the EPRI competition framework and describe key characteristics of the problem including; operational constraints, reward formulation, state representation and the economic dispatch approaches used.

### 3.1. Benchmark and data

We use the EPRI AI-ccelerating Unit Commitment competition dataset [26], a single-bus representation of the Irish test system also used by Za'ter et al. [9] and Fritz et al. [10], with instances generated in UnitCommitment.jl [51]. It contains 2,621 operating instances labelled 2021–2029 across all four quarters, the count also reported by Fritz et al. [10]. Each instance defines a 72-hour horizon at hourly resolution, with demand, wind and solar profiles, the initial status of each unit, and the technical and cost characteristics of 51 thermal generators with a combined capacity of 8,515 MW. The unit data are identical across instances; only the profiles and initial conditions vary. Twenty-four medium units (100–500 MW) provide 82.9% of capacity. Production cost curves are convex and piecewise linear with one to four segments, start-up costs reach about $46,000, and several units with high start-up costs have low marginal costs, which creates the usual trade-off between cycling and running.

We restrict the problem to the thermal fleet, the dominant flexible resource and the source of every binary commitment decision. This is the simplest version of the problem that retains its combinatorial core, which makes the behaviour of the learned policies easier to diagnose: the hydro, battery and pumped-storage assets of the full competition system would add continuous decisions that carry state between hours and could absorb or mask the effects studied here. Extending the analysis to the full system is left to future work (Section 7). Perfect foresight of demand, wind and solar is assumed, consistent with the competition formulation, so results characterise deterministic scheduling. The training set comprises 300 instances drawn at random from all quarters except the fourth quarters of 2026–2029; the validation set comprises 25 instances from those fourth quarters; and 100 test instances were drawn at random from the instances not used for training or validation. Each instance also carries the benchmark solution of the full competition system.

### 3.2. Unit commitment formulation

The thermal-only problem minimises production, start-up and penalty cost over the horizon,

$$\min_{u,s,w,p} \sum_{t=1}^{T} \left[ \sum_{i \in \mathcal{G}} \left( C_i(p_{i,t}) + C_i^{\mathrm{su}}\, s_{i,t} \right) + \lambda^{\mathrm{uns}} E_t^{\mathrm{uns}} + \lambda^{\mathrm{sp}} E_t^{\mathrm{sp}} \right] \qquad (1)$$

where $C_i(p_{i,t})$ is zero when the unit is offline and otherwise the no-load cost at minimum output plus the convex piecewise-linear cost of output above it, modelled with segment variables bounded by $u_{i,t}$. The objective is subject to demand balance, output limits, and start-up and shut-down logic,

$$\sum_{i} p_{i,t} + E_t^{\mathrm{uns}} - E_t^{\mathrm{sp}} = \max(0, D_t^{\mathrm{net}})$$

$$\underline{P}_i u_{i,t} \le p_{i,t} \le \overline{P}_i u_{i,t}$$

$$u_{i,t} - u_{i,t-1} = s_{i,t} - w_{i,t}$$

and to ramp limits between consecutive hours, with a start-up ceiling of $\max\left(R_i^{\mathrm{up}}, \underline{P}_i\right)$, and to minimum up- and down-time constraints including the initial conditions. The unserved-energy penalty is the instance's power balance penalty, $\lambda^{\mathrm{uns}} = 12{,}533$ \$/MWh, and the spill penalty is $\lambda^{\text{sp}} = 500$ \$/MWh. We solve Eq. (1) with HiGHS [52], warm-started from the merit-order rule of Section 4.2. Its constraint set mirrors the RL environment exactly, including the shut-down rules of Section 3.3, so that its solution is the optimum of the problem the agents face. The benchmark solution supplied with each instance solves a different problem, the full competition system with hydro and storage, and is used only as the denominator of the optimality ratio.

### 3.3. Markov decision process

Commitment is formulated as a finite-horizon episodic Markov decision process with one episode per instance and one decision per hour. At each step the agent proposes a binary vector $\mathbf{a}_t \in \{0,1\}^{51}$, where $a_{i,t} = 1$ requests unit $i$ to be online.

**Constraint handling.** A unit that has not completed its minimum up-time cannot be switched off, and a unit that has not completed its minimum down-time cannot be switched on. A unit whose minimum output exceeds its ramp-down limit cannot be switched off at all, which applies to two units. The benchmark's shutdown limits are otherwise not modelled, so other units may shut down from any output level (Section 6.4). A unit that none of these rules binds is termed free. Two ways of enforcing the rules are compared. In the standard PPO agents, they are enforced after sampling: any request that violates them is overridden, so the action that is executed can differ from the action the agent chose and is updated on. All other agents mask invalid actions before sampling [37], so that every sampled action is executed as chosen.

**Reward.** The reward at step $t$ is the negative normalised operating cost,

$$r_t = -\frac{F_t + S_t + \lambda^{\text{uns}} E_t^{\text{uns}} + \lambda^{\text{sp}} E_t^{\text{sp}} - B_t}{10^5}, \qquad (2)$$

where $S_t$ is start-up cost and $B_t$ is an optional baseline. Two formulations of Eq. (2) are used in the main experiments (Table 1). In the shaped reward R2, unserved energy and spill are computed from committed capacity, $E_t^{\text{uns}} = \max\left(0, D_t^{\text{net}} - \sum_i u_{i,t}\,\overline{P}_i\right)$ and $E_t^{\text{sp}} = \max\left(0, \sum_i u_{i,t}\,\underline{P}_i - D_t^{\text{net}}\right)$, the spill penalty is a design parameter (1,000 or 3,000 \$/MWh), and two shaping terms penalise units committed beyond 115% of net demand (\$100 per unit-hour) and over-commitment ahead of falling demand (500 \$/MWh); its fuel term is the cost of the dispatched energy, computed under proportional loading or, in some fine-tuning runs, economic dispatch (Section 3.5, Table 7). Two further shaped rewards, R0 and R1, which omit the fuel term or charge it at each committed unit's minimum output, gave similar results and are reported in Appendix C. The *clean* reward equals the MILP objective of Eq. (1), with unserved energy and spill of the actual dispatch, so that ramp and start-up limits are priced, and no shaping terms. In the *relative* variant, $B_t$ is the clean cost of the MILP schedule at the same instance and hour; it depends only on the instance and hour, so it does not change the optimal policy.

**Table 1.** Reward formulations used in the main experiments; R0 and R1 are defined in Appendix C.

| Reward | Fuel term $F_t$ | Unserved and spill | $\lambda^{\text{sp}}$ (\$/MWh) | Shaping terms | Dispatch |
|---|---|---|---|---|---|

| Reward | Fuel term $F_t$ | Unserved and spill | $\lambda^{sp}$ ($/MWh) | Shaping terms | Dispatch |
|---|---|---|---|---|---|
| R2 | $\sum_i C_i\left(p_{i,t}\right)$ | from committed capacity | 1,000 / 3,000 | yes | proportional or economic |
| Clean | $\sum_i C_i\left(p_{i,t}\right)$ | from actual dispatch | 500 | no | economic or look-ahead |

### 3.4. State representations

Two nested observation designs are used (Table 2): a basic design, and a balance-aware design that also makes the consequences of the current commitment, such as whether the fleet is obliged to spill, directly observable. Both include constraint clocks, which make the observation Markov with respect to minimum up- and down-times, and the full 72-hour net-demand forecast, so every agent has the same information about future demand as the MILP. All continuous variables are normalised.

**Table 2.** Observation designs.

| Feature group | Description | Basic | Balance-aware |
|---|---|---|---|
| Time encoding | sin/cos of hour of day and remaining-horizon fraction | • | • |
| Net-demand forecast | sliding window of the next 72 hours, and minimum, maximum, mean and slope over the next 8 hours | • | • |
| Commitment status | on/off with time-to-can-start and time-to-can-stop clocks | • | • |
| Balance indicators | committed $\sum \underline{P}_i$ and $\sum \overline{P}_i$ as ratios of net demand | | • |
| Imbalance feedback | previous-hour spill and unserved energy, normalised | | • |
| Cumulative starts | start-ups per unit so far in the episode | | • |
| Total dimension | | 232 | 287 |

### 3.5. Economic dispatch

Economic dispatch is resolved after commitment, following the standard two-stage decomposition in which the agent addresses the combinatorial commitment problem and a deterministic method allocates output [1,19,22]. Three methods are used, each respecting ramp limits relative to the previous hour's dispatch. *Proportional loading*, used with R2 unless stated, sets each unit at minimum output and distributes the residual in proportion to committed headroom, approximating practice in which reserve is spread across the fleet [53], and snaps outputs to discrete levels of 5% of $\overline{P}_i$. *Economic dispatch* fills the residual from the cheapest cost-curve segments of the committed units, which is exact within each hour for convex piecewise-linear costs and reproduces the MILP dispatch cost for a given commitment to within 0.01%. Both are myopic: the agent observes the full forecast but controls only commitment, and an hourly dispatcher cannot raise output ahead of a steep rise in net demand. *Look-ahead dispatch* solves a linear programme over the next four hours with the current commitment held fixed, the known net demand and the penalties of Eq. (1), and applies the first hour. On the MILP's own schedules for the 100 test instances, the hourly economic dispatcher leaves 5.8 MWh of unserved energy per instance and is feasible on 80% of instances; the look-ahead dispatcher leaves less than 0.01 MWh and is feasible on all of them. These comparisons use the MILP's schedules on the test instances, not those of any learned policy, so they do not favour any of the policies compared. A four-hour window performed better than six hours, because a longer window extends the assumption that the current commitment persists; the dispatcher is therefore an approximation to the joint commitment and dispatch of the MILP. All results in Section 5 use look-ahead dispatch unless stated.

## 4. Methodology

In this section we provide an overview of the methodology behind our experiments. We start in Sections 4.1 and 4.2 with a description of our experimental pipeline and the reference policies used for comparison. We then describe the imitation and training protocols used as well as the evaluation framework. Section 4.6 describes the warm-start experiment, in which the agent's schedule is used to warm-start the MILP solver and to fix some of its variables, and Section 4.7 concludes with a summary of the experiments.

### 4.1. Agents

Figure 1 summarises the experimental pipeline. *PPO* [54] is implemented in Stable-Baselines3 [55] over a MultiBinary(51) action space, mapping the observation to 51 independent Bernoulli probabilities, with generalised advantage estimation (GAE) [56]; the masked variant uses MaskablePPO from the same library. *DQN* [57] uses the action-branching architecture [33]: a shared network of three 256-unit ReLU layers with an off/on value pair per unit, a double-DQN target [58] averaged over branches, and masking of invalid actions. Its exploration flips each free unit's greedy decision with probability $\varepsilon$, the per-unit noise of Section 4.5. Hyperparameters are listed in Table 3.

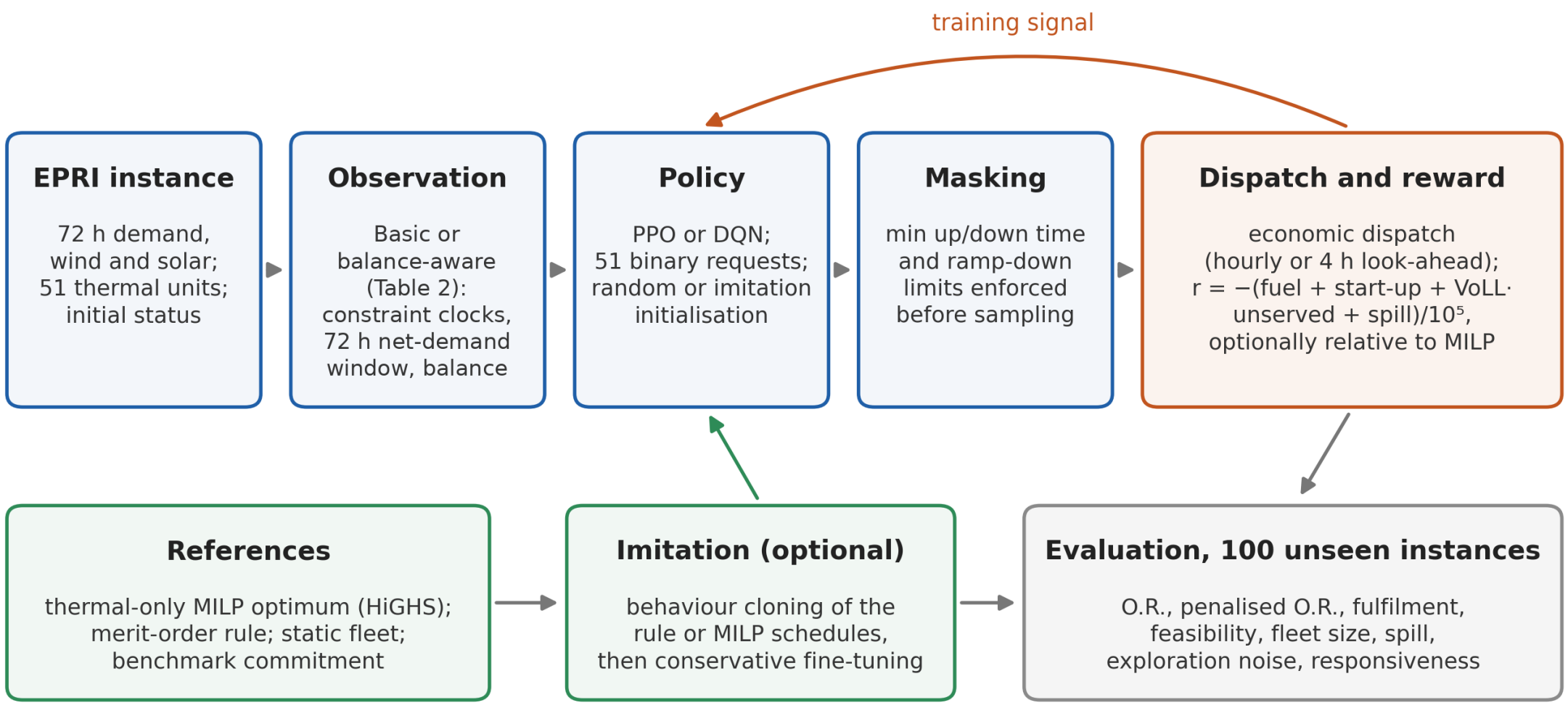

***Fig. 1.*** *Experimental pipeline. The agent observes the instance, proposes a commitment vector that is masked (or projected) onto legal transitions, and is trained on the negative operating cost of the resulting dispatch. Policies may be initialised by imitation of the merit-order rule or of the MILP solutions. All policies, the reference policies and the static-fleet control are dispatched and evaluated on the same 100 unseen instances.*

**Table 3.** Agent hyperparameters.

| Parameter | PPO (standard) | PPO (conservative fine-tuning) | DQN |
|---|---|---|---|
| Network | 2 × 64 (policy and value) | 2 × 64 | 3 × 256, 51 branches |
| Learning rate | $3 \times 10^{-4}$ | $1 \times 10^{-4}$ | $1 \times 10^{-4}$ (Adam) |
| Discount factor $\gamma$ | 0.99 | 0.99; 1.0 with the relative reward | 0.99 |

| Parameter | PPO (standard) | PPO (conservative fine-tuning) | DQN |
|---|---|---|---|
| Exploration | entropy coefficient 0.01 | entropy coefficient 0 | per-unit flips, $\varepsilon$: 0.5 → $\varepsilon_{\text{final}}$ over 150k steps, or constant |
| Rollout / batch | 256 steps, batch 64, 10 epochs | 8 × 32 steps, batch 64; 16 × 72 steps, batch 288 with the relative reward | replay 100,000; batch 256; update every 8 steps |
| Other | GAE $\lambda = 0.95$; clip range 0.2 | clip range 0.1 | soft target $\tau = 0.005$; Huber loss; gradient-norm clip 10 |

### 4.2. Reference policies

Four reference policies are evaluated with the same dispatch and cost pipeline as the agents.

**Thermal-only MILP optimum.** Eq. (1) is solved for all 100 test instances with a 100 s limit and a 0.1% relative gap on one CPU core; 77 were solved to proven optimality, the largest remaining gap was 0.46% and the median solve time was 52 s. We refer to this solution as the optimum; where optimality was not proven, it lies within 0.46% of the bound. The same model was solved for all 300 training and 25 validation instances with a 40 s limit (median gap 0.11%, 95th percentile 0.78%), to provide demonstrations and relative-reward baselines. Its commitment is replayed through the environment and the evaluation dispatcher, so that its results include any effect of the dispatcher.

**Merit-order rule.** At each hour, the units the environment forces online are kept, and further units are added in order of full-load average cost, $C_i(\overline{P}_i)/\overline{P}_i$, skipping units locked offline, until committed capacity reaches $m$ times net demand. The margin $m \in \{1.0, 1.05, 1.1, 1.2\}$ is the rule's only parameter.

**Static-fleet control.** To bound what any adaptation contributes, the $k$ cheapest units are committed for all 72 hours of every instance, for $k$ from 8 to 51 and under both full-load and first-segment cost orderings. The lower envelope of optimality ratio against fulfilment over all fleets defines the frontier attainable without any scheduling behaviour; policies are compared with it at equal fulfilment by linear interpolation.

**Benchmark commitment.** The thermal commitment of the benchmark solution is replayed through the thermal-only environment. Because the benchmark also uses hydro and storage, its thermal commitment does not meet the thermal-only demand in full.

### 4.3. Imitation and fine-tuning

For imitation, the schedules of a teacher (the merit-order rule or the MILP) on the 300 training instances are converted into state–action pairs under the agent's own observation. For the rule, a further 300 episodes are generated in which 10% of free units are flipped at random when executed but labelled with the rule's decision, so that the policy also sees states off the rule's path; for the MILP this is not possible without re-solving from each perturbed state, so only its own trajectories are used. The policy network is trained by masked cross-entropy for 100 epochs, and the value network is pre-trained on the discounted returns of the teacher's trajectories (or set to zero for the relative reward). For DQN, the branching network is first trained by masked cross-entropy and then by temporal-difference learning with a large-margin loss on the teacher's decisions [45], and the replay buffer is seeded with the teacher's transitions. A pre-trained policy that is not fine-tuned is termed imitation-only. The pre-trained policies are then fine-tuned online with the settings in Table 3.

### 4.4. Training protocol

Each PPO configuration is trained for $10^6$ environment steps (about 13,900 episodes) with three seeds (42, 43 and 44), except six of the seven tuning variants of Section 4.7 (all but the zero-entropy variant), which use one seed. DQN is trained for $5 \times 10^5$ steps from scratch with one seed per exploration rate, and for $3 \times 10^5$ steps from a pre-trained network with three seeds. Checkpoints are evaluated every 25,000–100,000 steps on the 25 validation instances, and the checkpoint with the lowest median penalised optimality ratio (Section 4.5) on validation is reported on the 100 test instances. The exceptions, reported at their final checkpoint, are DQN trained from scratch and the standard fine-tuning runs B1 and B3 of Section 5.4, whose drift is the quantity of interest. Training runs on a single CPU core per run and takes about 15–20 min for $10^6$ PPO steps, excluding validation evaluations. A trained policy produces a 72-hour schedule in about 0.07 s, or 0.12 s including look-ahead dispatch, against a median MILP solve time of 52 s. Results are reported as the mean and standard deviation over seeds of the per-seed statistics. The code uses Python 3.10, Stable-Baselines3 2.4.1 with sb3-contrib 2.4.0 and PyTorch 2.5.1, and HiGHS 1.15.1, and was run on a four-core Linux virtual machine (aarch64).

### 4.5. Evaluation metrics

**Optimality ratio.** Following the competition, the optimality ratio is

$$\mathrm{O.R.} = \frac{\sum_t \sum_i \left(C_i(p_{i,t}) + C_i^{\mathrm{su}} s_{i,t}\right)_{\mathrm{policy}}}{\sum_t \sum_i \left(C_i(p_{i,t}) + C_i^{\mathrm{su}} s_{i,t}\right)_{\mathrm{benchmark}}}, \qquad (3)$$

computed per instance with the thermal production and commitment of the benchmark solution in the denominator, and summarised by the median over instances. Unserved-energy and spill penalties are excluded from the numerator for comparability with the benchmark. Because the benchmark also uses hydro and storage, an O.R. of 1 is not attainable by a thermal-only schedule; we therefore also report each policy's mean per-instance gap to the thermal-only MILP optimum over the same 100 instances.

**Penalised optimality ratio.** The numerator of Eq. (3) is replaced by the full objective of Eq. (1), including unserved energy and spill priced at $\lambda^{\mathrm{uns}}$ and $\lambda^{\mathrm{sp}}$. Spill includes renewable surplus in hours of negative net demand, which no thermal schedule can avoid and which is common to all policies.

**Fulfilment and feasibility.** Demand fulfilment is the share of positive net demand served over the horizon. An instance is feasible if its total unserved energy is below 1 MWh, and the feasibility rate is the share of feasible instances.

**Fleet size and responsiveness.** Mean committed units $N_t$ is reported for every policy. Demand responsiveness $\rho$ is the within-episode Pearson correlation of $N_t$ with net demand, averaged over instances.

**Exploration noise.** For a stochastic policy, the per-unit noise at a state is the mean, over units free to change state, of $\min(\pi_i, 1 - \pi_i)$, where $\pi_i$ is the probability of committing unit $i$: the probability that the sampled decision differs from the greedy one. It is averaged over the states visited by the greedy policy on 20 test instances. For DQN it equals $\varepsilon$ by construction.

### 4.6. Learned policies as solver warm starts

A policy's commitment schedule is passed to the MILP of Eq. (1) in two ways. As a warm start, the schedule, completed by the dispatch for that commitment, is given to HiGHS as an initial solution. With variable fixing, the commitment of unit-hours with a policy commitment probability below $\tau \in \{0.001, 0.02, 0.05\}$ is also fixed at zero, following Fritz et al. [10]; decisions to commit are not fixed, because in a pilot on the first 20 of the 100 test instances fixing them cost 1.6% on average; the choice applies equally to every policy, so it does not favour any of them. The warm starts compared are no warm start, the merit-order rule ($m = 1.1$), the imitation-only policy of the MILP, the final fine-tuned PPO configuration and the default masked PPO agent trained from scratch; fixing is applied with three seeds of the imitation-only policy (two further seeds were trained for this purpose) and of the fine-tuned PPO. Two further controls test whether the gains need a learned policy at all. A heuristic fixes every unit-hour left off by the merit-order rule with a generous margin $m \in \{1.3, 1.5, 2.0\}$, warm-started from that rule's schedule. To separate how many variables a policy fixes from which ones it chooses, the imitation-only policy is also made to fix, on each instance, exactly as many unit-hours as the fine-tuned policy of the same seed index at $\tau = 0.001$ and $0.02$, choosing those it is most confident should be offline. Each solve uses one thread, a 100 s limit and a 0.1% relative gap, on all 100 test instances. Quality is the cost above the best solution found for the instance by any unrestricted solve, and speed-up is the ratio to the solve time without warm start on the same instance and machine (solves were split over two machines by instance); it is a lower bound, because that solve reached the time limit on 42% of instances.

### 4.7. Experiments

Seven experiments are reported (Table 4). **E1** evaluates the reference policies. **E2** trains standard PPO under R0, R1 and R2 with the basic observation (spill penalty 1,000 \$/MWh) and the balance-aware observation (3,000 \$/MWh), and DQN from scratch with final exploration rates of 0.02, 0.05 and 0.1. **E3** trains masked PPO (balance-aware, 3,000 \$/MWh, R2) with the default settings, with seven variants (observation and reward normalisation, a 256 × 256 network with eight parallel environments, a lower learning rate, entropy coefficients of 0 and 0.03, a clip range of 0.1, and learning-rate decay), and with the clean reward. **E4** measures the agents' exploration noise, simulates the reward-optimal fleet under noise, compares the agents with the rule under matched noise, and fine-tunes a lean policy with injected noise. **E5** fine-tunes policies that imitate the merit-order rule: PPO with standard (B1), conservative (B2) and standard settings without an entropy bonus (B3), and DQN with constant exploration. **E6** fine-tunes conservatively from imitation of the rule or the MILP under progressively refined rewards and dispatch. **E7** uses the policies as warm starts and variable-fixing guides for the MILP (Section 4.6).

**Table 4.** Experiments, the question each addresses, and where the results are reported.

| Experiment | Question | Results |
|---|---|---|
| E1 | How close do simple reference policies come to the optimum? | Section 5.1; Table 5 |
| E2 | Do agents trained from scratch learn lean schedules, across rewards and observations? | Section 5.2; Table 6, Fig. 2 |
| E3 | Do masking, tuning or a reward without shaping terms change the outcome? | Section 5.2; Table 6 |

| Experiment | Question | Results |
|---|---|---|
| E4 | Is over-commitment a response to exploration noise? | Section 5.3; Figs. 3 and 4, Table D1 |
| E5 | Does a lean policy drift towards over-commitment when fine-tuned? | Section 5.4; Fig. 5, Table 7 |
| E6 | Do imitation and conservative fine-tuning close the gap? | Section 5.5; Table 7 |
| E7 | Can the learned policy accelerate an exact solver? | Section 5.6; Table 8, Fig. 6 |

## 5. Results

This section first establishes the reference points (Section 5.1), then examines the over-commitment of agents trained from scratch and its link to exploration (Sections 5.2–5.4). Sections 5.5 and 5.6 report the imitation-based remedies and the use of the resulting policy within the MILP, and Section 5.7 the sensitivity of reported results to evaluation choices.

Results are reported mainly as two gaps to the thermal-only MILP optimum on the same instances: the gap in production and start-up cost, and the penalised gap, which also prices unserved energy and spill. A policy can score well on the first by leaving demand unserved, so the two should be read together. The O.R. against the competition benchmark is given for comparability with other work; because the benchmark also uses hydro and storage, the optimum itself scores 1.090 on it.

### 5.1. Reference points

A one-parameter merit-order rule comes close to the optimum on cost but not on reliability, and any policy that adapts commitment to demand has a large advantage over a fixed fleet (Table 5). The thermal-only MILP optimum attains a median O.R. of 1.090 with full fulfilment and 12.9 committed units on average. It does not reach 1.000 because the benchmark serves about 7% of energy with hydro and storage: its own thermal commitment, replayed here, gives 1.020 but serves only 93.4% of demand. A standalone thermal scheduler should therefore be judged against 1.090, not 1.0.

**Table 5.** Reference policies on the 100 unseen test instances under look-ahead dispatch. O.R. is the median over instances. Gaps to the MILP optimum are means of per-instance ratios over the same 100 instances; the penalised gap prices unserved energy and spill. Feasible is the share of instances with less than 1 MWh unserved. Bold marks the lowest gaps among the rules and static fleets.

| Policy | Gap to MILP | Penalised gap | O.R. | Fulfilment | Feasible | Mean fleet | $\bar{\rho}$ |
|---|---|---|---|---|---|---|---|
| Thermal-only MILP optimum | 0 | 0 | 1.090 | 100.00% | 100% | 12.9 | 0.972 |
| Merit-order rule, $m = 1.0$ | **+1.6%** | +24.3% | 1.108 | 99.84% | 19% | 12.7 | 0.981 |
| Merit-order rule, $m = 1.05$ | +2.8% | +14.1% | 1.120 | 99.91% | 52% | 13.3 | 0.982 |
| Merit-order rule, $m = 1.1$ | +4.1% | +11.1% | 1.137 | 99.95% | 58% | 13.8 | 0.982 |
| Merit-order rule, $m = 1.2$ | +7.1% | **+10.1%** | 1.166 | 99.98% | 66% | 15.1 | 0.981 |
| Static fleet, $k = 26$ | +63.2% | +198% | 1.558 | 99.95% | 92% | 26.0 | – |
| Static fleet, $k = 28$ | +67.0% | +202% | 1.589 | 99.98% | 93% | 28.0 | – |
| Benchmark commitment | −7.4% | +1,192% | 1.020 | 93.44% | 0% | 11.9 | 0.972 |

The rule comes within 1.6–7.1% of the optimum on the O.R., with fleets of 12.7–15.1 units. The margin trades cost against reliability: the tightest rule is cheapest on fuel and start-up but leaves 215 MWh unserved per instance, so on the penalised ratio the order reverses and the most generous margin is best (+10.1%). Fixed fleets are much more expensive. A static fleet must cover the peak in every hour, so it needs 26–28 units for the 99.95–99.98% fulfilment that the rules achieve, at an O.R. of 1.56–1.59 (Table B1). Any policy that adjusts commitment to demand therefore has a large advantage over a fixed fleet. Responsiveness does not distinguish the rule from the optimum (0.97–0.98); fleet size is the more informative diagnostic.

**5.2. Over-commitment in agents trained from scratch**

Agents trained from scratch converged to about twice the optimal fleet, whatever their reward, observation or settings (Table 6, Fig. 2). Under R2, both standard PPO configurations held 27–29 units and were 47–51% above the optimum. Neither the observation design nor the fuel term changed this: the balance-aware observation and the alternative rewards R0 and R1 each moved the mean fleet by at most four units (Appendix C).

**Table 6.** Agents trained from scratch for $10^6$ steps (DQN: $5 \times 10^5$), look-ahead dispatch, 100 unseen instances; mean ± SD over three seeds unless stated. Configurations are given as reward, observation and spill penalty; results under R0 and R1 are in Table C1. The default masked agent was selected on validation under proportional loading, as it was trained; the agents trained on the clean reward, and the default agent in the row below it, were selected under look-ahead dispatch. The frontier gap compares each policy with the static-fleet frontier at the same fulfilment. Bold marks the lowest gaps, excluding rows that report ranges.

| Agent | Gap to MILP | Penalised gap | O.R. | Fulfilment | Mean fleet | $\bar{\rho}$ | Frontier gap |
|---|---|---|---|---|---|---|---|
| *Projected-action PPO (reward, observation, spill penalty in $/MWh)* | | | | | | | |
| R2, basic, 1,000 | +47.0 ± 1.2% | +87 ± 20% | 1.539 ± 0.019 | 99.76 ± 0.12% | 28.8 ± 0.3 | 0.845 | +7.5% |
| R2, balance-aware, 3,000 | +51.2 ± 1.9% | +112 ± 47% | 1.568 ± 0.009 | 99.39 ± 0.64% | 27.4 ± 0.4 | 0.838 | +11.9% |
| *Masked PPO* | | | | | | | |
| Default: R2, balance-aware, 3,000 | **+40.9 ± 3.3%** | +55 ± 9% | 1.470 ± 0.050 | 99.92 ± 0.05% | 25.6 ± 1.6 | 0.847 | −1.7% |
| Default, look-ahead selection | +46.2 ± 2.6% | +57 ± 7% | 1.540 ± 0.029 | 99.92 ± 0.07% | 27.5 ± 2.0 | 0.861 | +1.7% |
| Clean reward, economic training | +46.4 ± 2.0% | +52 ± 2% | 1.525 ± 0.012 | 99.98 ± 0.02% | 29.3 ± 0.9 | 0.859 | −3.2% |
| Clean reward, look-ahead training | +43.5 ± 2.4% | **+48 ± 3%** | 1.499 ± 0.027 | 99.99 ± 0.01% | 27.6 ± 0.7 | 0.874 | −6.3% |
| Seven tuning variants[a] | +43 to +61% | +63 to +113% | 1.52–1.64 | 99.54–99.92% | 26.5–28.9 | 0.72–0.87 | +4.9 to +13.5% |
| *DQN* | | | | | | | |
| Clean reward, $\varepsilon_{\text{final}} = 0.02$–0.1[b] | +39 to +55% | +176 to +705% | 1.45–1.61 | 96.73–99.39% | 25.1–29.9 | 0.60–0.75 | – |

[a] One seed per variant, except the variant with entropy coefficient 0 (three seeds: O.R. 1.524 ± 0.008, 26.5 ± 0.6 units). [b] One seed per value of $\varepsilon_{\text{final}}$; final checkpoint after $5 \times 10^5$ steps.

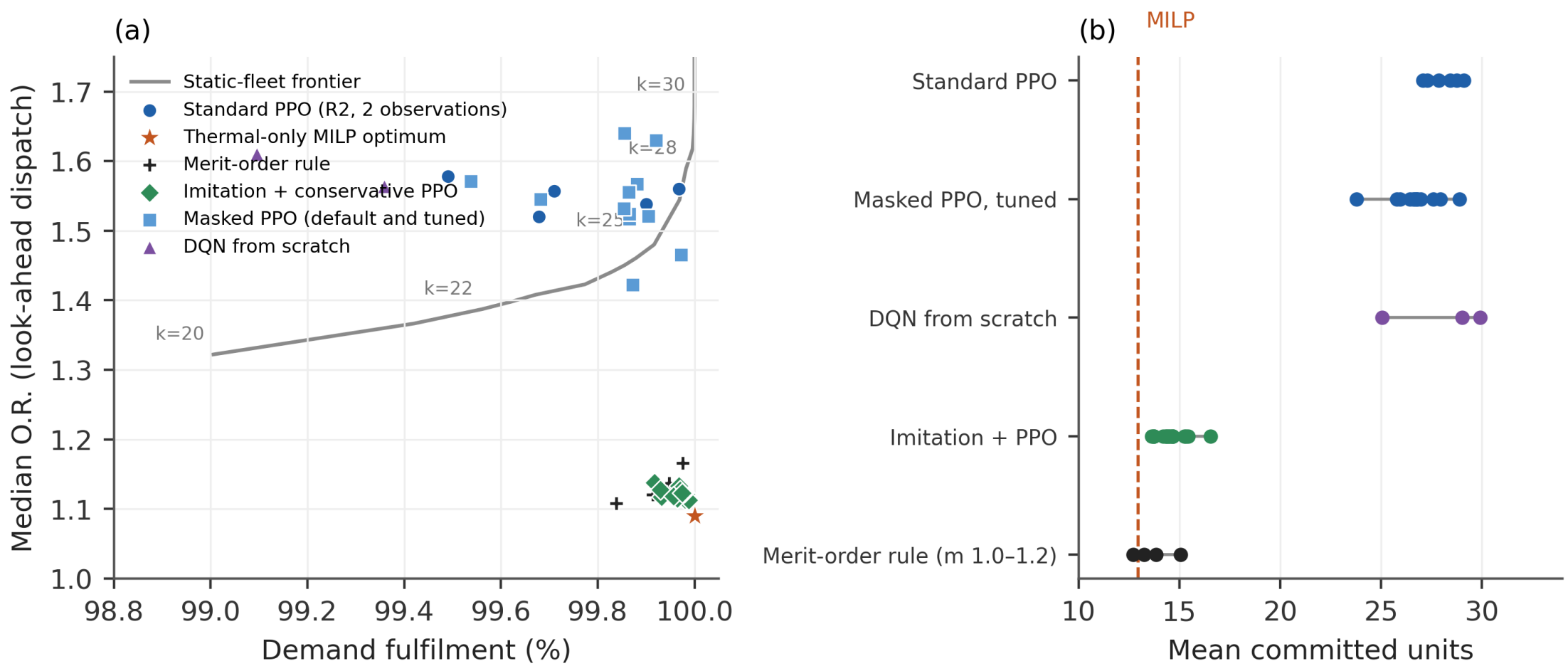

***Fig. 2.*** *Agents trained from scratch against the reference policies. (a) Median O.R. against demand fulfilment, under look-ahead dispatch, with the static-fleet frontier (lower envelope over fleets of the k cheapest units). Each marker is one trained policy (one seed). (b) Mean committed units; each marker is one policy, and the dashed line is the MILP optimum. Standard PPO is shown under R2; R0 and R1 are reported in Appendix C.*

Invalid-action masking was the only intervention that helped, removing the 70–110 overrides per episode of projected actions and bringing the fleet to 25.6 units, at the static-fleet frontier but still 40.9% above the optimum. None of the seven tuning variants improved on it, and DQN trained from scratch behaved similarly (25–30 units, 39–55% above the optimum), with no ordering of fleet by final exploration rate.

Removing the shaping terms did not change this either. Trained on the clean reward, which equals the MILP objective and has no shaping terms, the default masked agent held 27.6–29.3 units and was 43.5–46.4% above the optimum, depending on whether economic or look-ahead dispatch was used in training (Table 6). It was more reliable than under R2, meeting the feasibility threshold on about 90% of instances against 71%, but no leaner, and its exploration noise was unchanged at 0.29. The clean reward requires one of these dispatchers, so the comparison with R2, trained under proportional loading, also changes the training dispatch; selected under look-ahead validation like the clean-reward agents, the R2 agent held 27.5 units and was 46.2% above the optimum.

Executed greedily, the agents also performed poorly on their own training reward: the merit-order rule outperformed every agent under R2, by a factor of about 1.6 to 5. Section 5.3 shows that this ranking reverses under the agents' training noise.

The over-committed policies are demand-responsive ($\bar{\rho}$ = 0.72–0.87), but around a level about twice too high, and their fleets of 25–32 units lie close to the 25.5 units that a randomly initialised factorised policy commits in expectation. This coincidence admits a simpler reading, that the agents move little from where they start. However, Fig. A1 shows that the fleets fall from their initial level during the first $10^5$ steps and then remain there, and Section 5.4 shows that a lean policy drifts up towards a similar level, which that reading does not predict.

### 5.3. Exploration noise as insurance

Why would an agent settle on a policy its reward rates poorly? During training, each unit's decision is sampled independently, so committed units are switched off at random. Switching off a needed unit risks unserved energy priced at 12,533 \$/MWh, whereas keeping a spare unit online costs its no-load and minimum-output fuel, typically \$1,000–3,000 per hour. Under noise, spare capacity is insurance, and a policy that is optimal for the noisy behaviour policy over-commits when executed greedily.

We tested this by injecting the same kind of noise into the merit-order rule: at every hour, each free unit's decision was flipped with probability $\varepsilon$, and the rule was scored on a training reward over 50 training instances for a range of margins. Under R2, the reward-optimal fleet rises from 12.5 units without noise to 19.5 at $\varepsilon = 0.1$ and 26.2 at $\varepsilon = 0.3$. Under the clean reward, used for the DQN agents (Fig. 3a), it is 14.7 units without noise ($m = 1.2$) and 15.1, 16.9, 18.0, 22.4 and 27.8 units at $\varepsilon = 0.01$, 0.02, 0.05, 0.1 and 0.2. At high noise the cost is also nearly flat over a wide range of fleet sizes: at $\varepsilon = 0.1$, fleets of 19–28 units cost within about 5% (R2) to 7% (clean reward) of each other, so the reward gives little signal to shrink the fleet.

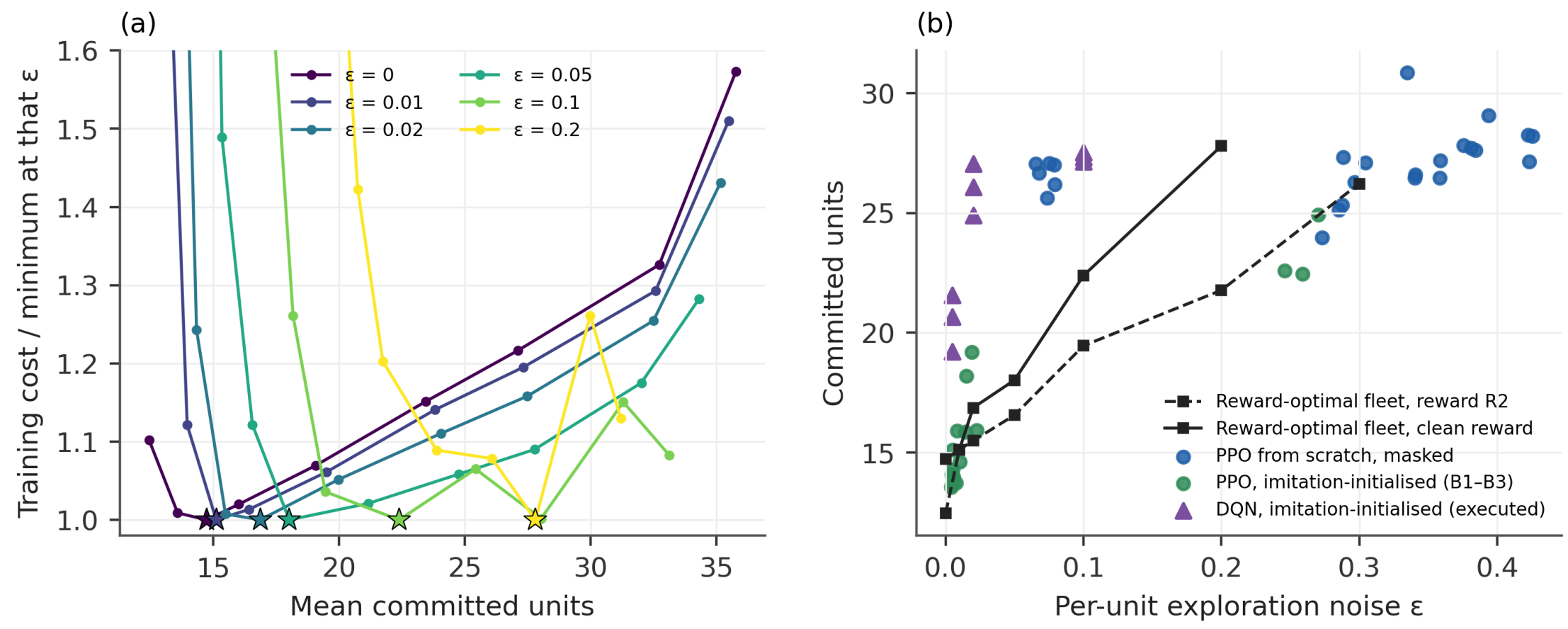

***Fig. 3.*** *Exploration noise as insurance. (a) Training cost (clean reward) of the merit-order rule with per-unit exploration noise ε injected, against the resulting mean fleet, for margins from 1.0 to 3.0; stars mark the reward-optimal fleet at each ε. (b) Reward-optimal fleet against ε under R2 and the clean reward, with the measured noise and greedy fleet of trained PPO agents (circles) and the noise and executed fleet of DQN agents initialised by imitation (triangles).*

Most trained agents lie close to the curve (Fig. 3b). The masked PPO agents of Section 5.2 had a per-unit noise of 0.27–0.43 and held 24–31 units; under R2 the reward-optimal fleet is 26.2 units at $\varepsilon = 0.3$, the highest noise simulated. Because the cost is nearly flat at these noise levels, this agreement is a weak test on its own, which motivates the direct test below. The agents' noise stays high because of the entropy bonus: the variant without it ended training with a noise of only 0.07. That variant does not fit the curve, however: it kept 26–27 units against a predicted 18–19. The imitation-initialised PPO agents of Section 5.4 lie near the curve at much lower noise; the DQN agents at the lowest noise lie above it (Section 5.4).

If over-commitment were simply a failure to optimise, the agents' stochastic policies should also lose to the rule when both act with the agents' training noise; they do not. On the R2 training reward over 30 training instances, the stochastic policies of all twelve masked agents cost 16–63% less than the best merit-order rule with the same average per-unit noise. Matching the noise separately for committed and uncommitted units gave the same result. The agents therefore perform well on the problem they face during training, which includes their own exploration, and poorly only when that policy is executed greedily. The comparison is approximate: the agents' noise is concentrated on units with commitment probabilities near one half, whereas the injected noise is spread uniformly and is costlier, so it favours the agents, and the simulated curve models their noise only approximately.

The account also makes a direct prediction: adding exploration noise to a lean policy should enlarge its fleet. We tested it by fine-tuning the final configuration of Section 5.5 from its imitation-only starting policy, with each free unit's executed decision flipped with probability ε during training only, as environment noise (Fig. 4, Table D1). The greedy fleet rose from 15.0 units without injected noise to 17.0–19.1 units at ε = 0.02–0.2, and the penalised gap from 4.5% to 13–62%. Up to ε = 0.05 the fleets lay within about one unit of the reward-optimal fleet of the noisy rule. At higher noise they fell well short of it (17.8 against 22.4 units at ε = 0.1, and 19.1 against 27.8 at ε = 0.2); at ε = 0.2 the fleet was still rising steeply when training ended, which suggests that the conservative step size limits how far the policy moves, but at ε = 0.1 it had nearly levelled off. Exploration noise therefore causes the fleet to grow, by about the predicted amount at low noise, though not to the predicted optimum at high noise within this budget.

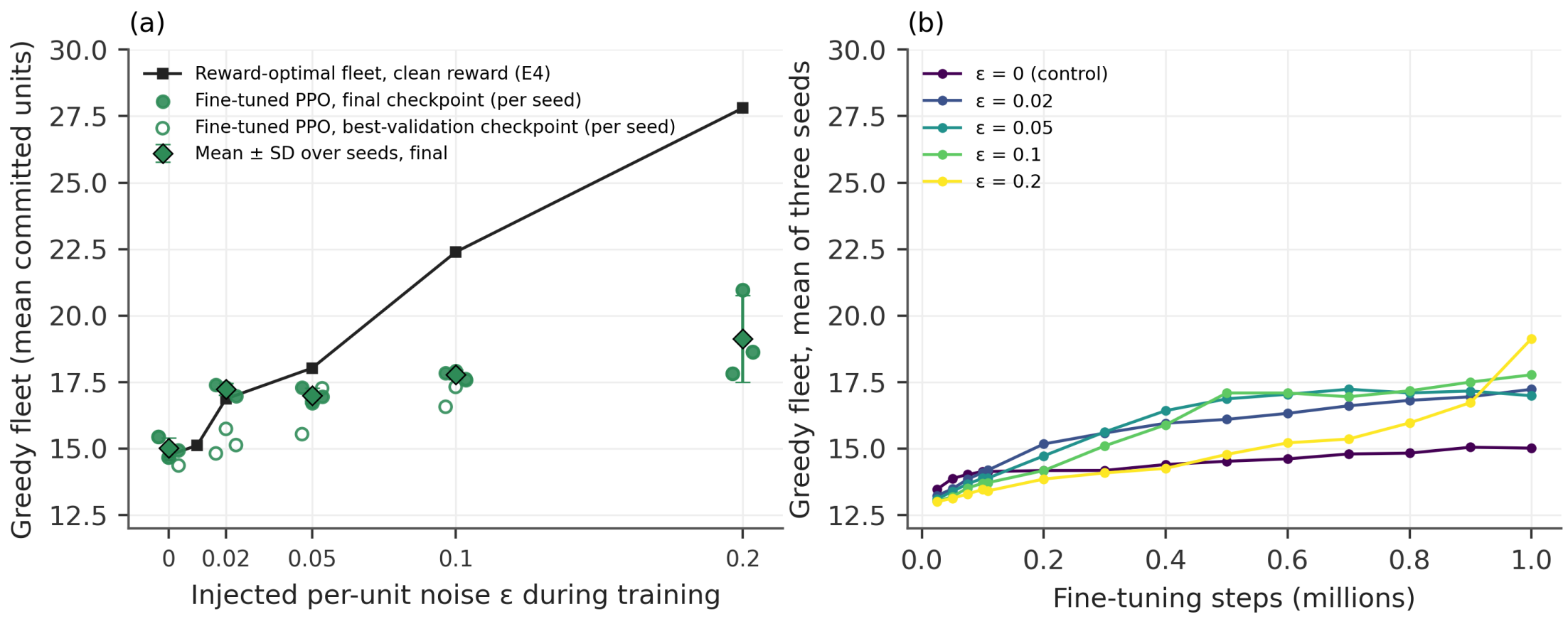

***Fig. 4.*** *Direct test of the insurance account. The final configuration of Section 5.5 fine-tuned from its imitation-only starting policy, with each free unit's executed decision flipped with probability ε during training. (a) Greedy fleet on the test instances against ε: final (filled) and best-validation (hollow) checkpoint of each seed, mean ± SD of the final checkpoints (diamonds), and the reward-optimal fleet of the noisy merit-order rule under the clean reward from Fig. 3 (squares). (b) Mean greedy fleet over three seeds during fine-tuning.*

### 5.4. Drift towards over-commitment from a lean start

If over-commitment is a response to exploration, a lean initial policy should drift towards it when it explores. We initialised PPO by imitation of the merit-order rule ($m = 1.1$) and fine-tuned it under R2 with three settings (Fig. 5a, Table 7). With the standard settings of Section 5.2 (B1), the fleet rose steadily from 13.8 to 22.6–24.9 units over $10^6$ steps, with a final per-unit noise of 0.25–0.27. Removing the entropy bonus but keeping the larger steps (B3) slowed the drift to 15.8–18.8 units after $10^6$ steps, with a noise of 0.008–0.019. With conservative settings (B2; Table 3), the fleet stayed at 13.6–14.3 units for the full $10^6$ steps, with noise of about 0.005, and the policy improved slightly on its teacher (Table 7). The B1 and B2 fleets lie close to the reward-optimal fleet at their measured noise (Fig. 3b); B3 lies 1–4 units above it. Because B2 changes the learning rate, the clip range and the entropy bonus together, it does not separate smaller steps from lower noise; B3 removes the entropy bonus alone and drifts less than B1.

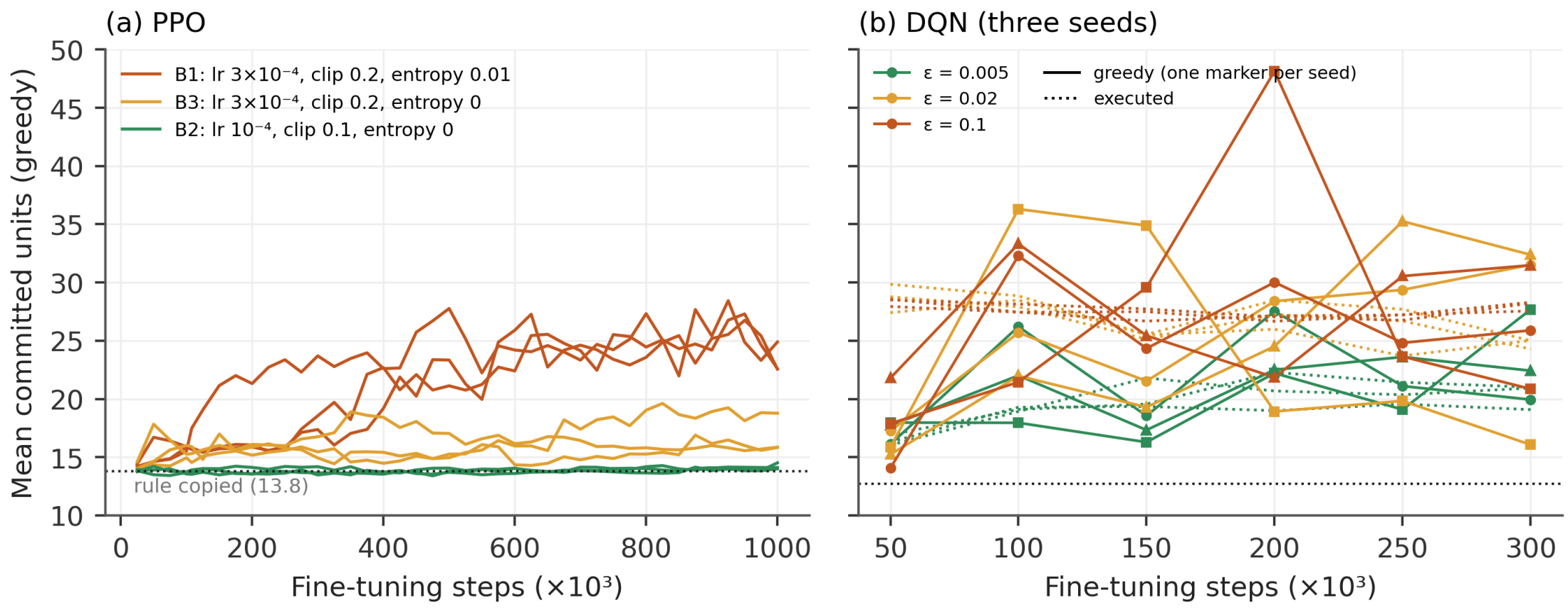

*__Fig. 5.__ Fleet size during fine-tuning from a policy that imitates the merit-order rule. (a) PPO with standard (B1), standard without entropy bonus (B3) and conservative (B2) settings; one line per seed. (b) DQN with constant exploration ε, one line per seed; solid lines are the greedy fleet on the test instances and dotted lines the fleet executed during training.*

The DQN experiment varied the noise directly. Initialised by imitation of the rule ($m = 1.0$) and fine-tuned with constant per-unit noise, its executed fleet over the last third of training increased with $\varepsilon$: 20.5, 26.0 and 27.3 units at $\varepsilon$ = 0.005, 0.02 and 0.1, an increase of about seven units, close to the predicted increase between the lowest and highest noise levels. The shape differs, however: most of the observed increase occurred between the two lowest levels, where the simulated curve predicts only about two units. Even the almost noise-free arm rose from 12.7 to about 19–22 units, and the greedy fleet varied between 14 and 48 units from one checkpoint to the next. DQN therefore shows two components: a noise-dependent increase consistent with the insurance account, and a noise-independent drift that places even the lowest-noise arm about five units above the predicted 15 units. The latter is consistent with relative overgeneralisation in independent learners [36] (Section 6.1).

### 5.5. Imitation and conservative fine-tuning

Imitation followed by conservative fine-tuning closed most of the gap to the optimum (Table 7). Imitation alone was cheap but unreliable. The imitation-only policies reproduced 98.9–99.6% of their teachers' unit decisions, but small errors in which units were committed produced shortfalls once the policy left the teacher's trajectory: the imitation-only policy of the MILP had the lowest O.R. of any policy serving more than 98% of demand because it served only 98.6%, leaving 2,453 MWh unserved per instance. Conservative fine-tuning repaired this within $10^5$ steps by adding one to two units, after which every conservatively fine-tuned policy served 99.94–99.98% of demand.

**Table 7.** Policies initialised by imitation, look-ahead dispatch, 100 unseen instances; mean ± SD over three seeds for fine-tuned policies; B1 and B3 are reported at their final checkpoint. Teacher and reward are given for each, with R2 (economic) denoting R2 under economic dispatch; conservative fine-tuning uses the settings in Table 3. Feasible is the share of instances with less than 1 MWh unserved, as in Table 5. Bold marks the lowest gaps among the fine-tuned policies.

| Policy | Teacher | Reward | Gap to MILP | Penalised gap | O.R. | Fulfilment | Feasible | Mean fleet |
|---|---|---|---|---|---|---|---|---|
| Imitation-only | rule, $m = 1.1$ | – | +3.8% | +18.8% | 1.130 | 99.89% | 45% | 13.8 |
| Imitation-only | rule, $m = 1.0$ | – | +1.0% | +103% | 1.103 | 99.41% | 0% | 12.6 |
| Imitation-only | MILP | – | −0.9% | +253% | 1.083 | 98.60% | 0% | 12.9 |
| B1: standard fine-tuning, final | rule, $m = 1.1$ | R2 | +30.9 ± 4.3% | +74 ± 24% | 1.402 ± 0.043 | 99.73 ± 0.16% | 32 ± 11% | 23.3 ± 1.3 |
| B3: standard, no entropy bonus, final | rule, $m = 1.1$ | R2 | +8.8 ± 4.2% | +20.4 ± 1.3% | 1.186 ± 0.053 | 99.92 ± 0.03% | 56 ± 13% | 16.8 ± 1.7 |
| B2: conservative | rule, $m = 1.1$ | R2 | +3.6 ± 0.3% | +6.9 ± 0.3% | 1.125 ± 0.002 | 99.97 ± 0.00% | 68 ± 4% | 13.8 ± 0.3 |
| Conservative | rule, $m = 1.0$ | R2 (economic) | +3.7 ± 0.5% | +6.7 ± 0.7% | 1.127 ± 0.006 | 99.97 ± 0.01% | 68 ± 8% | 14.0 ± 0.4 |
| Conservative | MILP | R2 (economic) | +4.1 ± 0.9% | +10.6 ± 3.5% | 1.131 ± 0.006 | 99.94 ± 0.03% | 64 ± 11% | 15.5 ± 1.0 |
| Conservative | MILP | R2 (economic), relative | +2.9 ± 0.2% | +8.9 ± 1.9% | 1.117 ± 0.002 | 99.95 ± 0.02% | 61 ± 3% | 14.5 ± 0.2 |
| Conservative | MILP | clean, relative | +2.9 ± 0.3% | **+5.1 ± 0.4%** | 1.118 ± 0.004 | 99.98 ± 0.00% | 78 ± 8% | 14.9 ± 0.3 |
| Conservative, look-ahead training | MILP | clean, relative | **+2.6 ± 0.05%** | +5.2 ± 0.6% | 1.113 ± 0.001 | 99.98 ± 0.00% | 71 ± 6% | 14.8 ± 0.6 |

Imitation moved PPO from 41–64% above the optimum to 3.8%; conservative fine-tuning preserved this (3.6%) and cut the penalised gap from 18.8% to 6.9%, whereas standard fine-tuning undid it. Refinements of the teacher, reward and dispatch each gave smaller gains (Table 7): a reward relative to the MILP cost stopped the fleet creeping upwards, the clean reward cut the penalised gap from 8.9% to 5.1%, and training with the look-ahead dispatcher brought all three seeds to a gap of 2.6%. The two clean-reward configurations have the lowest penalised cost of any policy evaluated (+5.1–5.2%, against +10.1% for the best merit-order rule) and are feasible on 71–78% of instances, against 19–66% for the rules. On the O.R. alone, only the tightest rule is cheaper (+1.6%), and only because it leaves 215 MWh unserved per instance.

The remaining gap has two sources. Compared with the MILP schedule on the same 100 instances, the final policies hold more spare capacity: their median committed capacity is 1.16–1.17 times net demand, against 1.03 for the MILP. The no-load cost of the extra commitment ($0.80–0.83 million per instance) outweighs its lower energy cost ($0.56–0.58 million), while start-up costs are almost equal (+$0.01–0.02 million). The policies also leave 20–31 MWh unserved per instance, from mistimed rather than insufficient commitments, the compounding-error behaviour that interactive imitation [44] is designed to correct.

### 5.6. Learned policies as solver warm starts

Used to fix variables, the fine-tuned policy made the MILP several times faster at a small cost; used only as a warm start, it made the result more reliable but not faster (Table 8, Fig. 6). As a warm start alone, no policy shortened the solve materially (speed-up 1.1–1.2), because solve time is dominated by closing the bound. Without one, HiGHS reached the time limit on 42% of instances and averaged 2.4% above the best solution; with the fine-tuned PPO schedule or the merit-order rule, it averaged 0.07% and 0.10%. The imitation-only policy was a poor warm start (1.5% on average, up to 45%), because its shortfalls make its schedule heavily penalised.

**Table 8.** Learned policies as warm starts for the thermal-only MILP, 100 unseen instances, one thread, 100 s limit. Fixing sets the commitment of unit-hours with a policy commitment probability below $\tau$ to zero. Speed-up is the geometric mean over instances of the ratio to the solve time without warm start (a lower bound); gaps are costs above the best solution found for the instance. Fixing rows are means over three seeds (± SD for the mean gap), with Worst the maximum over all seeds' instances; warm-start rows, the from-scratch fixing row and the heuristic rows use one policy. Matched-count rows fix, on each instance, as many unit-hours as the fine-tuned PPO policy at the stated threshold.

| Warm start | Fixing | Variables fixed | Speed-up | Mean gap | 95th percentile | Worst | Within 0.1% |
|---|---|---|---|---|---|---|---|
| *Warm start only* | | | | | | | |
| None | – | – | 1.0 | 2.37% | 14.1% | 53.8% | 72% |
| Merit-order rule, $m = 1.1$ | – | – | 1.1 | 0.10% | 0.22% | 2.6% | 81% |
| Imitation-only policy | – | – | 1.1 | 1.50% | 11.2% | 44.6% | 80% |
| Masked PPO from scratch | – | – | 1.2 | 0.12% | 0.24% | 6.3% | 84% |
| Fine-tuned PPO (final) | – | – | 1.1 | 0.07% | 0.20% | 2.9% | 86% |
| *Variable fixing* | | | | | | | |
| Imitation-only policy | $\tau = 0.001$ | 61% | 5.7 | 0.23 ± 0.20% | 0.40% | 27.9% | 60% |
| Imitation-only policy | $\tau = 0.02$ | 69% | 17.0 | 4.47 ± 1.24% | 22.4% | 78.1% | 33% |
| Imitation-only policy | $\tau = 0.05$ | 71% | 24.9 | 13.5 ± 0.6% | 60.0% | 191% | 14% |
| Fine-tuned PPO | $\tau = 0.001$ | 59% | 3.8 | 0.10 ± 0.03% | 0.37% | 1.1% | 69% |
| Fine-tuned PPO | $\tau = 0.02$ | 64% | 6.6 | 0.19 ± 0.03% | 0.54% | 4.5% | 49% |
| Fine-tuned PPO | $\tau = 0.05$ | 66% | 8.5 | 0.25 ± 0.03% | 0.69% | 5.5% | 38% |
| Masked PPO from scratch | $\tau = 0.02$ | 5% | 0.6 | 2.08% | 4.9% | 13.6% | 0% |
| *Controls* | | | | | | | |
| Imitation-only, matched count | as PPO, $\tau = 0.001$ | 59% | 4.9 | 0.15 ± 0.07% | 0.37% | 8.9% | 63% |
| Imitation-only, matched count | as PPO, $\tau = 0.02$ | 64% | 8.8 | 0.60 ± 0.22% | 1.70% | 27.9% | 52% |
| Merit-order heuristic | $m = 2.0$ | 45% | 4.0 | 2.86% | 13.7% | 26.1% | 30% |
| Merit-order heuristic | $m = 1.5$ | 62% | 8.1 | 3.46% | 18.7% | 30.3% | 18% |
| Merit-order heuristic | $m = 1.3$ | 68% | 13.4 | 4.67% | 27.4% | 37.1% | 7% |

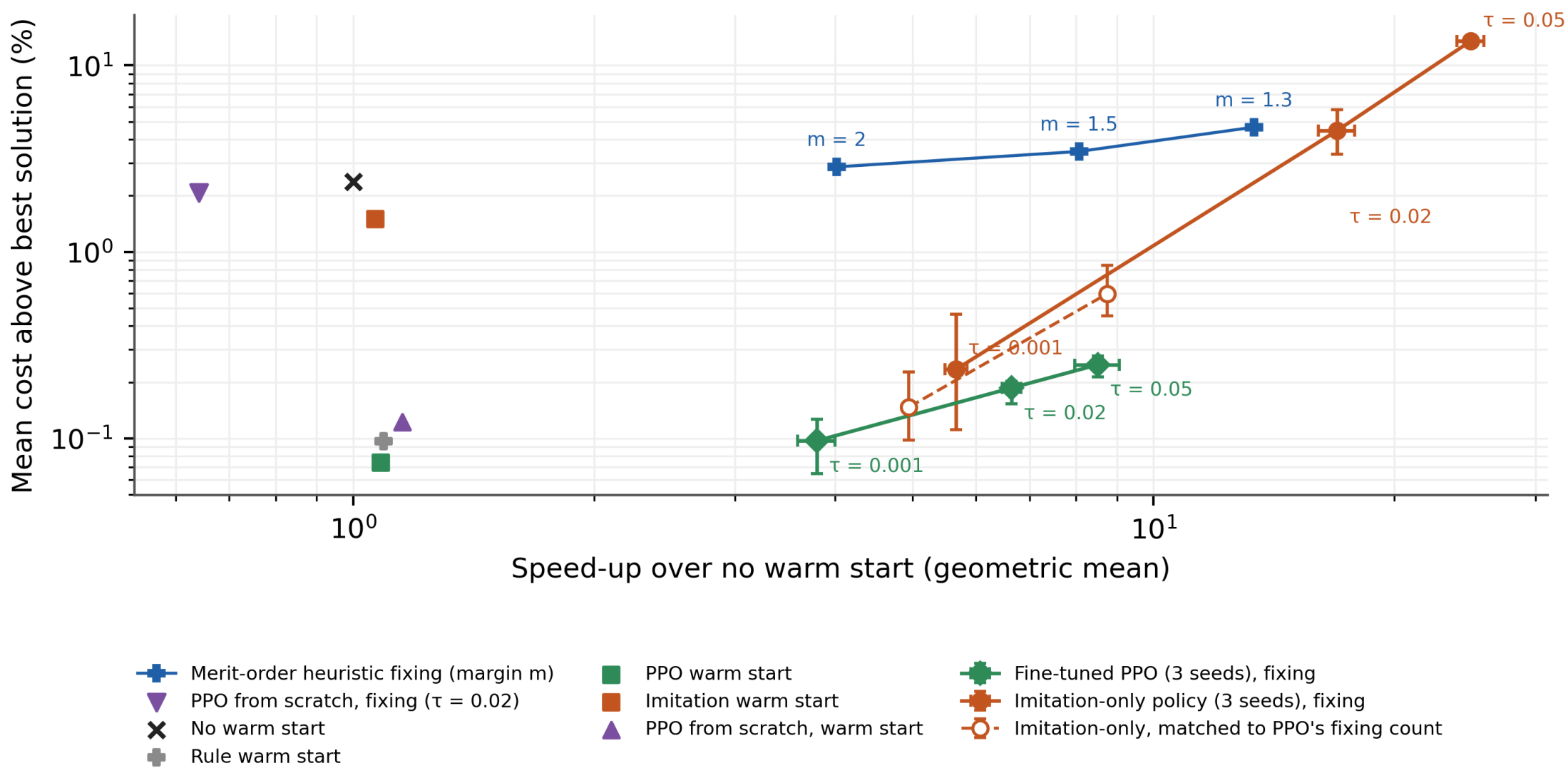

***Fig. 6.*** *Learned policies as solver warm starts. Cost above the best solution found against the speed-up over the solve without warm start, as the mean over 100 test instances. Solid lines connect fixing thresholds $\tau$ for the fine-tuned PPO and the imitation-only policy of the MILP (markers: mean over three seeds; bars: range over seeds); the dashed line is the imitation-only policy fixing as many unit-hours as PPO, and the blue line the merit-order heuristic at margins $m$. Single markers are warm starts without fixing and, for the agent trained from scratch, fixing at $\tau = 0.02$.*

Fixing the confident decisions of the fine-tuned policy gave a controlled trade-off: 3.8 times faster at 0.10% above the best solution, 6.6 times at 0.19% and 8.5 times at 0.25%, with a seed-to-seed standard deviation of 0.03 percentage points. The imitation-only policy fixed more variables at each threshold and was faster, but its confident decisions were less reliable: at $\tau = 0.001$ it was 5.7 times faster at 0.23% on average, and its mean cost rose steeply at higher thresholds (Table 8). At comparable speed-ups (5.7 and 6.6 times), the worst case was 27.9% for the imitation-only policy against 4.5% for the fine-tuned policy. The imitation-only policy was trained only to reproduce the MILP's decisions and can be confident in states its teacher never visited; fine-tuning on the cost of its own decisions is consistent with making the policy's confidence a better guide to which units can stay offline.

Two controls support the view that the learned policy matters. The merit-order heuristic was 4–13 times faster but 2.9–4.7% more expensive on average, with worst cases of 26–37%, so a good schedule alone is not enough to choose which variables to fix.

With matched counts (Section 4.6), the imitation-only policy remained faster. At the lower level it was indistinguishable in two of three seeds, with one seed reaching 8.9% on an instance; at the higher level its mean cost was three times larger (0.60% against 0.19%) and its worst case was 15.7–27.9% in every seed, against 0.9–4.5% for the fine-tuned policy. The two policies therefore trade speed against reliability: the imitation-only policy's choice of variables makes the restricted problem easier, while the fine-tuned policy's choice makes it safer, with the difference showing mainly in the tail.

The over-committing agent trained from scratch was confident about only 5% of decisions, and fixing those made the solve slower and 2.1% more expensive.

The speed-ups reported by Fritz et al. [10] (Section 2.1) were obtained on the full system, with a different solver and per-generator thresholds tuned on validation; the speed-ups here use a single threshold for all units, and the comparison is between policies within one pipeline.

### 5.7. Sensitivity to evaluation choices

The dispatch method used for evaluation changes the reported O.R. by amounts comparable to the differences between methods. Benchmark scope and the exclusion of unserved energy from the O.R. have similar effects, discussed in Sections 5.1 and 5.5.

Proportional loading spreads load across all committed units, including expensive ones. It raised the O.R. of the MILP commitment from 1.090 under look-ahead dispatch to 1.130 (+3.6%), and that of the rules by 0.9–3.0%, so it also changes the gaps between policies. Its discrete output levels leave shortfalls even when capacity is ample (459 MWh per instance on the MILP's own schedules), so that neither the MILP schedule nor any rule met the 1 MWh feasibility threshold on any instance. Hourly economic dispatch gives the same O.R. as look-ahead dispatch but lower feasibility (Section 3.5). Feasibility rates should therefore be reported with the dispatcher.

## 6. Discussion

This section draws the results together. Section 6.1 discusses why factorised agents over-commit, Section 6.2 the implications for learned schedulers, Section 6.3 a recommended reporting protocol, and Section 6.4 the limitations of the study.

Fig. 7 summarises our results visually, placing the main policies by their mean fleet and penalised cost relative to the optimum. Agents trained from scratch form one cluster at about twice the optimal fleet and 1.5–8 times the optimal penalised cost, whatever their algorithm, reward or settings. The merit-order rules and the conservatively fine-tuned policies sit close to the optimum, the latter within 5–14%, while standard fine-tuning moves a lean policy back towards the over-committed cluster.

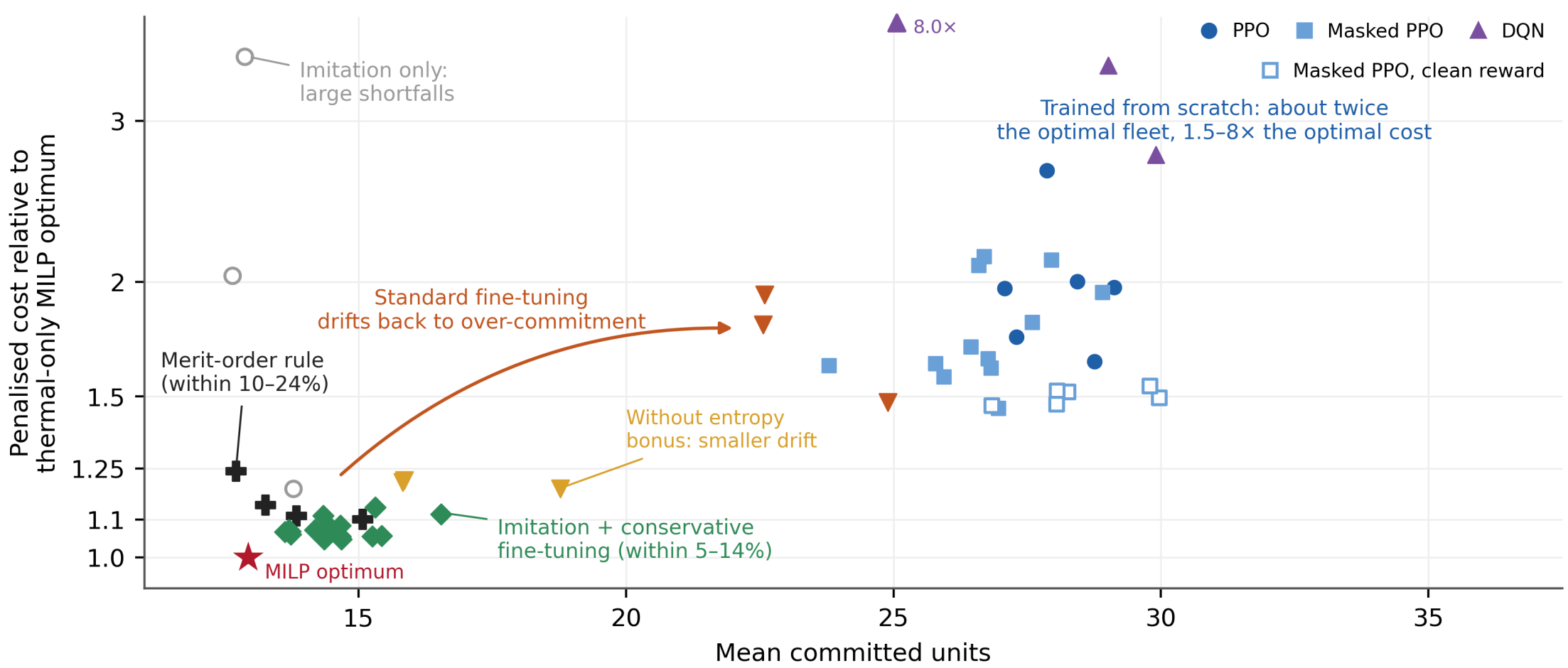

*Fig. 7. Summary of the main policies evaluated: penalised cost relative to the thermal-only MILP optimum (log scale) against mean committed units, on the 100 test instances under look-ahead dispatch. Each marker is one trained policy (one seed) or reference policy.*

*Policies trained from scratch are shown in blue (PPO, masked PPO, with masked PPO trained on the clean reward shown as open squares) and purple (DQN; the run at 8.0 times the optimum is drawn at the upper edge); the imitation-only policies are hollow; standard fine-tuning (B1) and fine-tuning without an entropy bonus (B3) are shown at their final checkpoints.*

### 6.1. Why factorised agents over-commit

The results support a specific mechanism. With a factorised action space, the policy optimised during training is a noisy one in which each unit's decision is sampled independently, and for that noisy policy spare commitment is a good answer: it insures against random de-commitments, which the unserved-energy penalty prices heavily. Evaluated greedily, the same policy over-commits. Four results support this account: injected exploration noise enlarged a lean policy's fleet, as exploration did for DQN; a reward without shaping terms left the over-commitment in place; under their own noise the agents outperformed the rule; and most PPO fleets lay close to the reward-optimal fleet at their measured noise. The fleet at high injected noise and the zero-entropy variant fit less well (Section 5.3). Joint or autoregressive action heads, which would remove the factorisation itself, were not tested.

The account is incomplete for DQN, whose fleet also drifted upwards with almost no noise and whose greedy policy was unstable between checkpoints. Independent value branches that each average over the others' behaviour favour actions that are safe individually over those that are jointly optimal [36], and heavy-tailed penalties make commitment such an action; we state this as a hypothesis. More generally, the effect resembles the preference of on-policy methods for safe behaviour under exploration [11], and may arise wherever many binary decisions jointly secure a constraint whose violation is expensive.

### 6.2. Implications for learned schedulers

Four practical implications follow. First, a learned scheduler should be compared with a simple rule and with an optimum for exactly the problem it solves; on this benchmark a one-parameter rule is within 1.6–7% of the optimum and sets the minimum bar a learned scheduler should clear. Second, because the gap is largely a response to exploration, remedies that reduce the need for it are effective, as imitation followed by conservative fine-tuning showed (Section 5.5). These comparisons are made under perfect foresight, where the MILP solves exactly the problem being scored; the setting therefore favours mathematical programming and gives no credit for the sequential, reactive decision-making that motivates RL. Under forecast uncertainty some spare commitment acts as reserve, so the cost of the behaviour diagnosed here may be smaller, a hypothesis that the extension to uncertain conditions in Section 7 would test. Third, the training reward and dispatch should match the evaluation objective.

Fourth, an RL policy and an exact solver are complementary. As a warm start, the fine-tuned policy made the MILP result more reliable, as did the merit-order rule; as a guide for variable fixing it gave a several-fold speed-up at a cost within a few tenths of a percent, with a smaller worst case than an imitation-only policy fixing as many variables. An over-committing policy trained from scratch is of little use for fixing; for hybrid pipelines that warm-start or repair a learned proposal [9,10,21], the results suggest that the calibration of the policy's confidence matters as much as its accuracy, which makes a conservatively fine-tuned RL policy a natural partner for an exact solver.

### 6.3. Recommended reporting protocol

On the basis of these results, we recommend that evaluations of learned UC schedulers report the following:

1. the optimality ratio with its distribution across instances, and a penalised ratio that prices unserved energy and spill;
2. demand fulfilment and feasibility rate adjacent to the ratio, with the dispatch method stated;
3. the gap to an optimum solved for exactly the problem the agent faces, where the benchmark's scope differs;
4. a simple rule, such as a merit-order rule with a capacity margin, evaluated with the same dispatch and cost pipeline;
5. mean committed fleet size, spill and unserved energy, and, for stochastic policies, the exploration noise at the reported checkpoint;
6. the agent's performance on its own training reward relative to the rule;
7. all metrics across at least three, and preferably five or more, independently seeded training runs [24], with checkpoints selected on validation instances.

Items 4–6 can be computed retrospectively for any study that saved its trained policies.

### 6.4. Limitations

Several limitations bound these conclusions.

1. *Deterministic setting.* All results assume perfect foresight of demand, wind and solar, so every policy is compared with a MILP that solves exactly the problem being scored. The study therefore does not measure how RL policies, or the over-commitment diagnosed here, perform under forecast uncertainty, where that optimum is not attainable and some spare commitment has value as reserve (Section 6.2).
2. *Optimum.* The model does not enforce the benchmark's shutdown limits, which would bind for 12 of the 51 units; enforcing them changes the optimum by less than 0.3% on average, but individual schedules would need adjusting to honour them.
3. *Warm-start study.* Solves used the open-source HiGHS solver on one thread, the thermal-only problem and a common threshold for all generators; speed-ups are lower bounds because the reference solve often reached its time limit, and the runs were split between two machines (each instance on one machine).

## 7. Conclusion

This first study of RL on the 51-unit EPRI benchmark found that PPO and DQN trained from scratch committed about twice as many units as a thermal-only MILP optimum. With factorised actions, spare units insure the agents against their own exploration: injecting exploration noise enlarged a lean policy's fleet, and removing the reward's shaping terms did not remove the over-commitment. Removing the need for that exploration closed most of the gap: imitation of the MILP solutions followed by conservative fine-tuning, with a reward equal to the MILP objective and look-ahead dispatch, came within 2.6% of the optimum and gave the lowest penalised cost of any policy evaluated. The same policy made an exact solver 3.8–8.5 times faster at a cost within 0.10–0.25% of the best solution, which suggests that RL policies for UC are most useful alongside, rather than instead of, mathematical programming.

Future work should address the remaining gap with interactive imitation, which relabels the states the policy visits with the expert's decisions [44], test joint or autoregressive action models, tune fixing thresholds per generator, and extend the analysis to the full system with storage, network constraints and forecast uncertainty.

## CRediT authorship contribution statement

**Maharani Devira Pramita:** Conceptualization, Methodology, Software, Formal analysis, Investigation, Data curation, Visualization, Writing – original draft. **Aidan O'Sullivan:** Conceptualization, Methodology, Software, Formal analysis, Supervision, Validation, Writing – review & editing.

## Declaration of competing interest

The authors declare that they have no known competing financial interests or personal relationships that could have appeared to influence the work reported in this paper.

## Funding

This research did not receive any specific grant from funding agencies in the public, commercial, or not-for-profit sectors.

## Data availability

The EPRI AI-ccelerating Unit Commitment dataset is available from the competition page [26]. The code for the environment, agents, MILP reference and evaluation pipeline, the trained model checkpoints and the result files underlying all tables are openly available under an MIT licence at https://github.com/devipramita/RL-UC-demand-responsiveness; the results reported here correspond to commit 9dfa996d.

## Declaration of generative AI and AI-assisted technologies in the writing process

During the preparation of this work the authors used Claude (Anthropic) in order to restructure and edit the manuscript text and format the references. After using this tool, the authors reviewed and edited the content as needed and take full responsibility for the content of the publication.

## Appendix A. Training curves

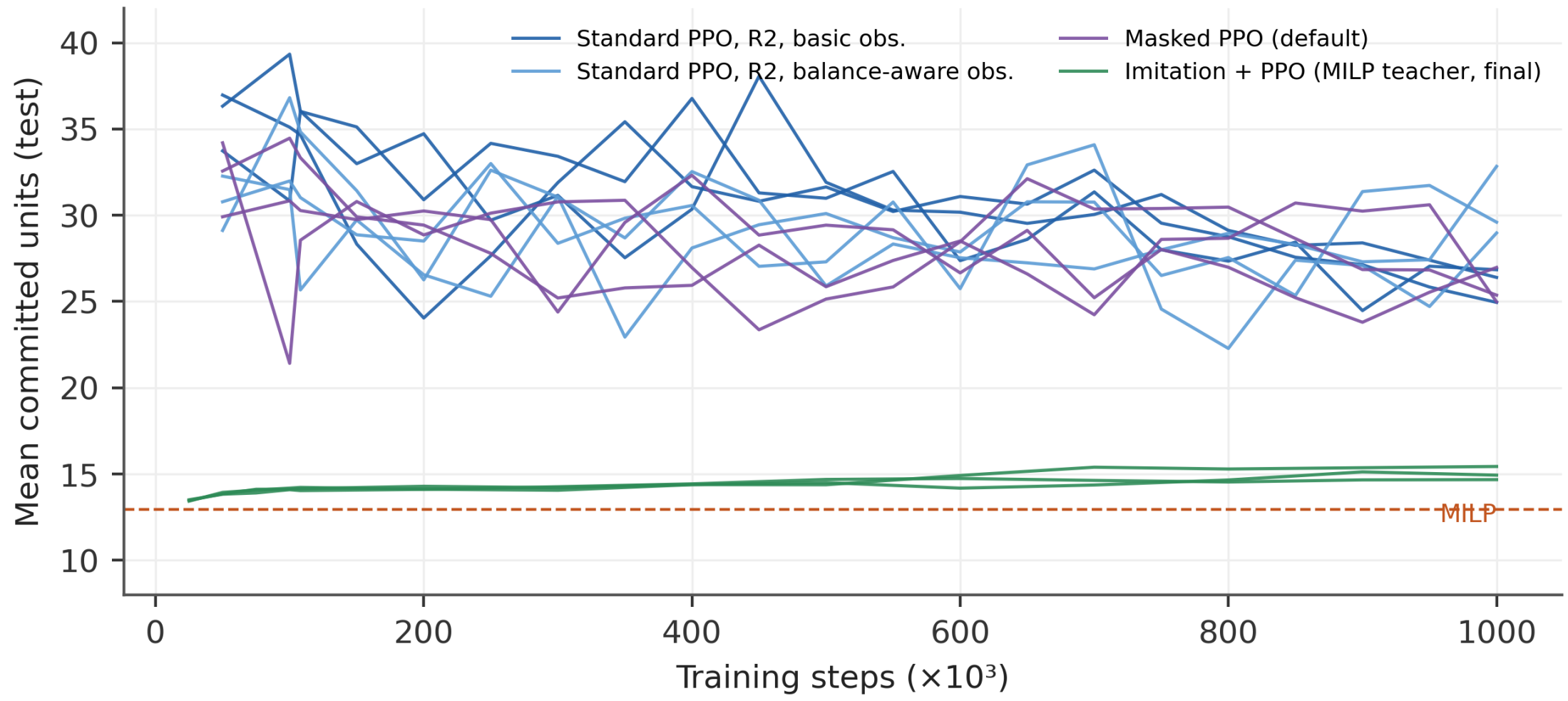

***Fig. A1.*** *Mean committed units on the test instances during training for standard PPO under R2 (basic and balance-aware observations), masked PPO and the final imitation-initialised configuration (one line per seed). The dashed line is the MILP optimum.*

## Appendix B. Static-fleet sweep

**Table B1.** Static-fleet control: the $k$ cheapest units by full-load average cost committed for all 72 hours, look-ahead dispatch, 100 unseen instances. The frontier in Fig. 2 is the lower envelope over this ordering and an ordering by first-segment marginal cost; results for the latter follow the same pattern and are available with the code.

| $k$ | O.R. 25th | O.R. median | O.R. 75th | Fulfilment | Spill (MWh) | Unserved (MWh) | Penalised O.R. |
|---|---|---|---|---|---|---|---|
| 10 | 0.458 | 0.553 | 0.738 | 56.9% | 10,748 | 94,712 | 109.9 |
| 15 | 0.966 | 1.091 | 1.386 | 91.9% | 20,815 | 18,943 | 20.79 |
| 18 | 1.067 | 1.221 | 1.531 | 96.5% | 26,285 | 8,315 | 8.94 |
| 20 | 1.150 | 1.322 | 1.666 | 99.01% | 30,898 | 2,271 | 3.97 |
| 22 | 1.196 | 1.387 | 1.760 | 99.56% | 33,915 | 988 | 3.46 |
| 24 | 1.212 | 1.408 | 1.788 | 99.67% | 34,385 | 737 | 3.10 |
| 25 | 1.265 | 1.484 | 1.911 | 99.88% | 37,568 | 278 | 2.91 |
| 26 | 1.316 | 1.558 | 2.036 | 99.95% | 40,958 | 108 | 2.82 |
| 28 | 1.341 | 1.589 | 2.090 | 99.98% | 42,064 | 38 | 2.89 |
| 30 | 1.405 | 1.680 | 2.242 | 100.00% | 45,851 | 0 | 3.13 |
| 51 | 1.678 | 2.066 | 2.845 | 100.00% | 58,099 | 0 | 4.14 |

## Appendix C. Additional shaped rewards

R0 and R1 share the structure of R2 (Section 3.3) and differ only in the fuel term: R0 omits it, and R1 charges each committed unit at its minimum output and first-segment marginal cost. Standard PPO was trained under both with the two configurations of Section 4.7, using action projection. Table C1 reports the results: the mean fleets differ from those under R2 (Table 6) by at most four units, and every configuration is 48–64% above the optimum.

**Table C1.** Standard PPO trained from scratch under rewards R0 and R1, for $10^6$ steps, look-ahead dispatch, 100 unseen instances; mean ± SD over three seeds. Columns as in Table 6.

| Agent | Gap to MILP | Penalised gap | O.R. | Fulfilment | Mean fleet | $\bar{\rho}$ | Frontier gap |
|---|---|---|---|---|---|---|---|
| R0, basic, 1,000 | +63.8 ± 6.5% | +125 ± 11% | 1.650 ± 0.028 | 99.82 ± 0.23% | 32.5 ± 3.2 | 0.758 | +11.0% |
| R0, balance-aware, 3,000 | +53.3 ± 4.6% | +101 ± 29% | 1.581 ± 0.031 | 99.82 ± 0.14% | 29.2 ± 0.8 | 0.867 | +7.3% |
| R1, basic, 1,000 | +48.6 ± 5.6% | +100 ± 25% | 1.551 ± 0.026 | 99.74 ± 0.21% | 29.8 ± 1.0 | 0.832 | +8.6% |
| R1, balance-aware, 3,000 | +49.2 ± 3.5% | +65 ± 2% | 1.556 ± 0.035 | 99.96 ± 0.04% | 28.7 ± 1.5 | 0.870 | +0.8% |

## Appendix D. Injected-noise fine-tuning

Table D1 reports the injected-noise experiment of Section 5.3 (E4). The control (ε = 0) reproduces the final configuration of Table 7, whose best-validation checkpoint is reported there. Checkpoints selected on validation are leaner than the final ones at every noise level below ε = 0.2, because validation selection favours the early checkpoints of a policy whose fleet is still growing; the final checkpoints are therefore the primary result, as for B1 and B3.

**Table D1.** Fine-tuning the final configuration with injected per-unit noise ε during training, look-ahead dispatch, 100 unseen instances; mean ± SD over three seeds at the final checkpoint, with the mean fleet at the best-validation checkpoint for comparison. The predicted fleet is the reward-optimal fleet of the noisy merit-order rule under the clean reward (Fig. 3). Own noise is the policy's per-unit noise (Section 4.5), excluding the injected noise.

| Injected ε | Gap to MILP | Penalised gap | O.R. | Fulfilment | Feasible (%) | Mean fleet | Fleet, best val. | Predicted fleet | Own noise |
|---|---|---|---|---|---|---|---|---|---|
| 0 (control) | +2.8 ± 0.2% | +4.5 ± 1.1% | 1.115 ± 0.002 | 99.99 ± 0.01% | 75 ± 10 | 15.0 ± 0.4 | 14.8 ± 0.6 | 14.7 | 0.025 ± 0.001 |
| 0.02 | +9.6 ± 1.6% | +13.2 ± 2.1% | 1.185 ± 0.015 | 99.98 ± 0.01% | 86 ± 9 | 17.2 ± 0.2 | 15.2 ± 0.5 | 16.9 | 0.032 ± 0.003 |
| 0.05 | +11.6 ± 0.6% | +17.3 ± 1.6% | 1.214 ± 0.011 | 99.97 ± 0.01% | 79 ± 6 | 17.0 ± 0.3 | 16.5 ± 0.9 | 18.0 | 0.038 ± 0.005 |

| Injected ε | Gap to MILP | Penalised gap | O.R. | Fulfilment | Feasible (%) | Mean fleet | Fleet, best val. | Predicted fleet | Own noise |
|---|---|---|---|---|---|---|---|---|---|
| 0.1 | +12.6 ± 1.3% | +33.2 ± 3.0% | 1.231 ± 0.014 | 99.87 ± 0.01% | 56 ± 3 | 17.8 ± 0.2 | 17.2 ± 0.5 | 22.4 | 0.045 ± 0.002 |
| 0.2 | +13.9 ± 3.3% | +62 ± 26% | 1.261 ± 0.037 | 99.68 ± 0.15% | 44 ± 13 | 19.1 ± 1.6 | 19.1 ± 1.6 | 27.8 | 0.069 ± 0.010 |